\documentclass[10pt,twocolumn,letterpaper]{article}

\usepackage[pagenumbers]{cvpr} 

\usepackage{graphicx}
\usepackage{booktabs}
\usepackage{amsmath,amssymb,graphicx}
\usepackage{stmaryrd}
\usepackage{amsfonts}
\usepackage{url}
\usepackage{algorithm}
\usepackage{algorithmicx}
\usepackage{algpseudocode}
\usepackage{makecell}
\usepackage{adjustbox}
\usepackage{multirow}
\usepackage{caption}
\usepackage{array}
\usepackage{color}
\usepackage{slashbox}
\usepackage{arydshln}
\newcommand{\RNum}[1]{\uppercase\expandafter{\romannumeral #1\relax}}

\graphicspath{{figure/}}

\newcommand{\keypoint}[1]{\vspace{0.0cm}\noindent\textbf{#1}\;}
\newcommand{\cut}[1]{}
\usepackage{ragged2e}
\renewcommand{\raggedright}{\leftskip=0pt \rightskip=0pt plus 0cm}
\usepackage[pagebackref,breaklinks,colorlinks]{hyperref}

\usepackage[capitalize]{cleveref}
\crefname{section}{Sec.}{Secs.}
\Crefname{section}{Section}{Sections}
\Crefname{table}{Table}{Tables}
\crefname{table}{Tab.}{Tabs.}

\DeclareMathOperator{\LCE}{Cross\_Entropy}
\DeclareMathOperator*{\Mlp}{Mlp}
\DeclareMathOperator*{\bert}{Bert}
\DeclareMathOperator*{\softmax}{Softmax}
\DeclareMathOperator*{\fc}{FC}
\DeclareMathOperator*{\sa}{SA}
\DeclareMathOperator*{\kl}{KL}
\DeclareMathOperator*{\mmd}{MMD}
\DeclareMathOperator*{\iou}{IoU}

\DeclareMathOperator*{\acc}{Acc}

\def\cvprPaperID{***} 
\def\confName{CVPR}
\def\confYear{2022}

\makeatletter
\def\thanks#1{\protected@xdef\@thanks{\@thanks
        \protect\footnotetext{#1}}}
\makeatother

\begin{document}

\title{Making a Bird AI Expert Work for You and Me}

\author{Dongliang~Chang,~Kaiyue~Pang,~Ruoyi~Du,~Zhanyu~Ma,~Yi-Zhe~Song, and~Jun~Guo  
\thanks{D. Chang, R. Du, Z. Ma, and J. Guo are with the Pattern Recognition and Intelligent
System Laboratory, School of Artificial Intelligence, Beijing University of Posts and Telecommunications, Beijing 100876, China (e-mail: mazhanyu@bupt.edu.cn).}
\thanks{
K. Pang and Y.-Z. Song  are with SketchX, CVSSP, University of Surrey, Guildford GU2 7XH, United Kingdom.}}

\maketitle

\begin{abstract}

As powerful as fine-grained visual classification (FGVC) is, responding your query with a bird name of ``Whip-poor-will" or ``Mallard" probably does not make much sense. This however commonly accepted in the literature, underlines a fundamental question interfacing AI and human -- what constitutes transferable knowledge for human to learn from AI? This paper sets out to answer this very question using FGVC as a test bed. Specifically, we envisage a scenario where a trained FGVC model (the AI expert) functions as a knowledge provider in enabling average people (you and me) to become better domain experts ourselves, i.e. those capable in distinguishing between ``Whip-poor-will" and ``Mallard".
Fig.~\ref{fig:intro-1} lays out our approach in answering this question. Assuming an AI expert trained using expert human labels, we ask (i) what is the best transferable knowledge we can extract from AI, and (ii) what is the most practical means to measure the gains in expertise given that knowledge? On the former, we propose to represent knowledge as highly discriminative visual regions that are expert-exclusive. For that, we devise a multi-stage learning framework, which starts with modelling visual attention of domain experts and novices before discriminatively distilling their differences to acquire the expert exclusive knowledge. For the latter, we simulate the evaluation process as book guide to best accommodate the learning practice of what is accustomed to humans. A comprehensive human study of 15,000 trials shows our method is able to consistently improve people of divergent bird expertise to recognise once unrecognisable birds. Interestingly, our approach also leads to improved conventional FGVC performance when the extracted knowledge defined is utilised as means to achieve discriminative localisation. Codes are available at: https://github.com/PRIS-CV/Making-a-Bird-AI-Expert-Work-for-You-and-Me
\end{abstract}

\section{Introduction}

\begin{figure}[t]
\begin{center}
  \includegraphics[width=0.95\linewidth]{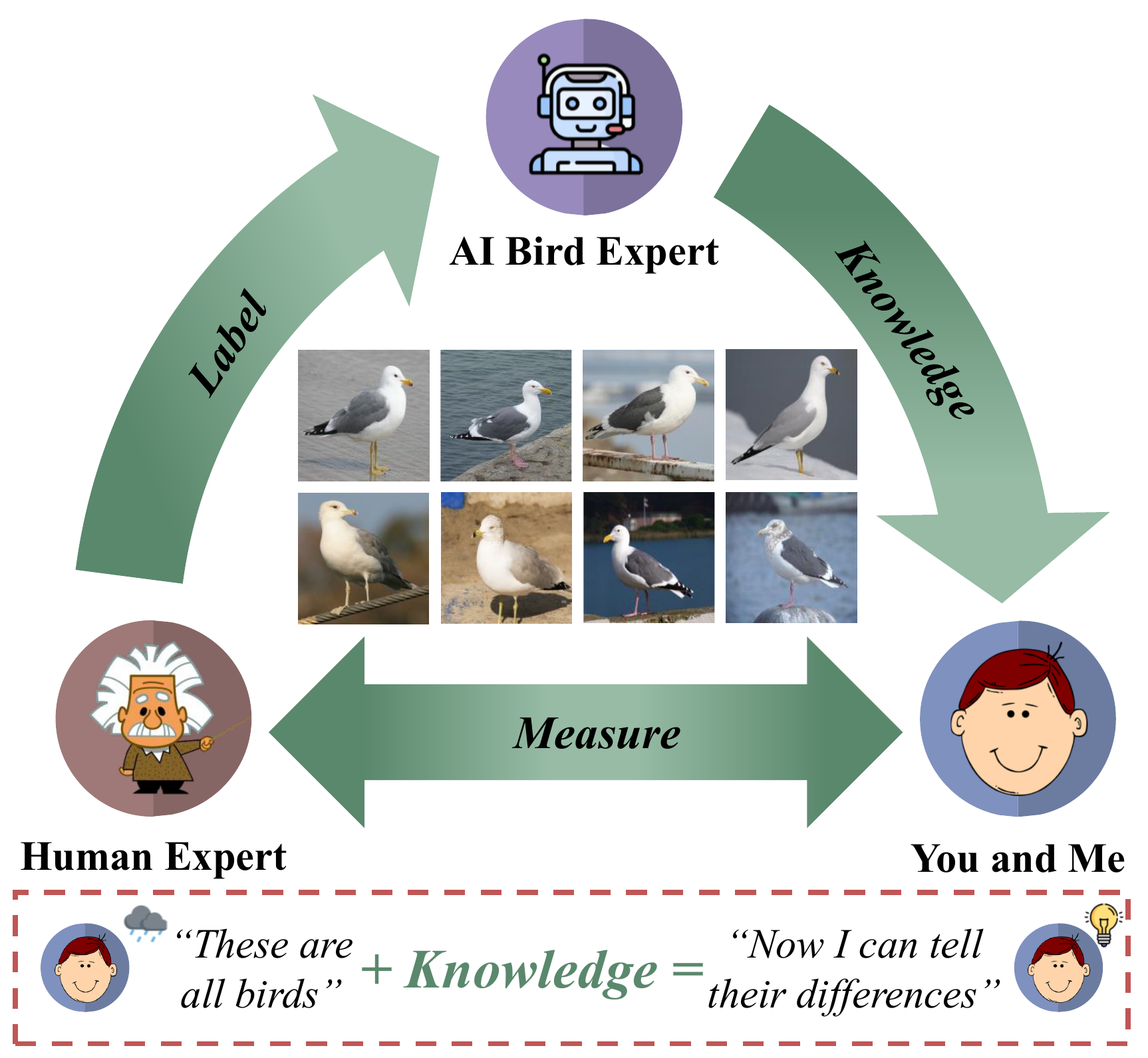}
\end{center}
\vspace{-0.2cm}
  \caption{AI Bird Expert Enriches Human Bird Knowledge. By retreating from the common goal of a FGVC model in pursuing better expert label predictions, we envision a human-centred FGVC endeavour and propose a first solution.}
\label{fig:intro-1}
\end{figure}

AI is great -- arguably the debate is on how it ultimately benefits mankind. Progress on computer vision has predominately followed the ``Human for AI" trend, where human data are used to train AI models that replace humans in some capacity. In this paper, we are interested in the complete opposite direction, \ie ``AI for Human", and ask the question ``can trained AI models help to enrich human knowledge instead?". 

We pick the problem of fine-grained visual classification (FGVC) as a test bed on this quest. FGVC is a good fit as one of the few areas in computer vision where AI agents\footnote{We call FGVC and AI Expert interchangeably throughout the paper.} can already reasonably \textit{replace} human experts \cite{biederman1999subordinate,deng2012hedging,berg2013poof,lin2015bilinear,zheng2017recognition,sun2018multi,yang2018learning,huang2020interpretable,joung2021learning}, \eg in identifying species of birds \cite{wah2011caltech}, models of cars and aircrafts \cite{krause20133d,maji13fine}, and tell one flower from another \cite{nilsback2008automated}. The question then becomes -- can the expert knowledge learned by AI be transferred across to an average human, so that ``you and me" become experts too? \ie those that can tell that the eight birds in Fig. \ref{fig:intro-1} are in fact from different species.

Fig.~\ref{fig:intro-1} illustrates the ambition of this paper -- to complete this three-way transfer cycle among human expert, AI expert and average human (you and me). The link where human experts provide labels to train an AI bird expert is the known part and precisely what FGVC in its conventional form strives for. Key for this paper is on how to make the remaining two connections: (i) how to extract \textit{knowledge} from AI that is digestible to a human (like a book), and (ii) how can we \textit{measure} the progress on ``you and me" becoming more expert-like using that \textit{knowledge}.

\begin{figure}[t]
\begin{center}
  \includegraphics[width=1.0\linewidth]{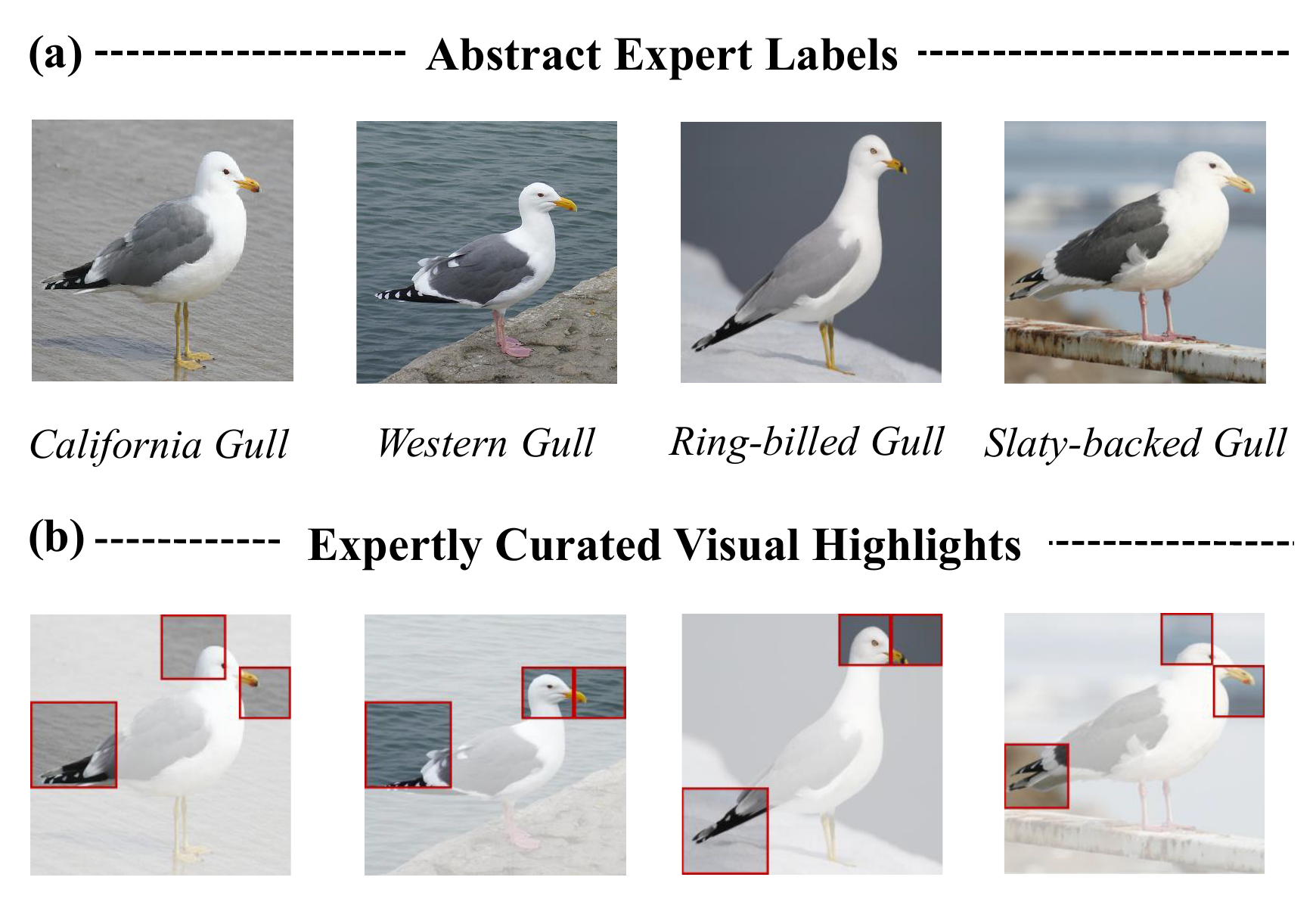}
\end{center}
\vspace{-0.3cm}
  \caption{Capitalising on knowledge in visual form (instead of abstract label inherent to a FGVC model), we show positive human feedback in digesting it towards better recognition.}
\label{fig:intro-2}
\end{figure}

On making the first link, we first stand with past works \cite{ordonez2013large,mac2018teaching,chang2021your} on the lack of interpretability of fine-grained expert labels to an average human (Fig. \ref{fig:intro-2}(a)). As such, they do not constitute good ``knowledge" in our context, \eg telling me the top left bird in Fig.~\ref{fig:intro-2} is a ``California Gull" probably does not say much. Our key innovation here is resorting to the highly discriminative regions that experts exclusively attend to as \textit{transferable} knowledge (Fig.~\ref{fig:intro-2}(b)). This echoes well with psychological findings on the importance of using visual highlights for novices to learn in complex visual tasks \cite{grant2003eye,roads2016using,hommel2019no}. 

To form the second connection and therefore close the loop, we literally take inspiration from a ``book" -- an expert bird guide in this context. More specifically, we present knowledge extracted from the AI expert as a bird guide to a human. The idea is then a \textit{better bird guide} (\ie knowledge) will result in ``you and me" becoming \textit{more expert-like}. We therefore take the degree to which the human has improved in being able to tell different species of birds as a measure of how good the extracted knowledge actually is. 

\begin{figure*}[t]
\begin{center}
  \includegraphics[width=0.95\linewidth]{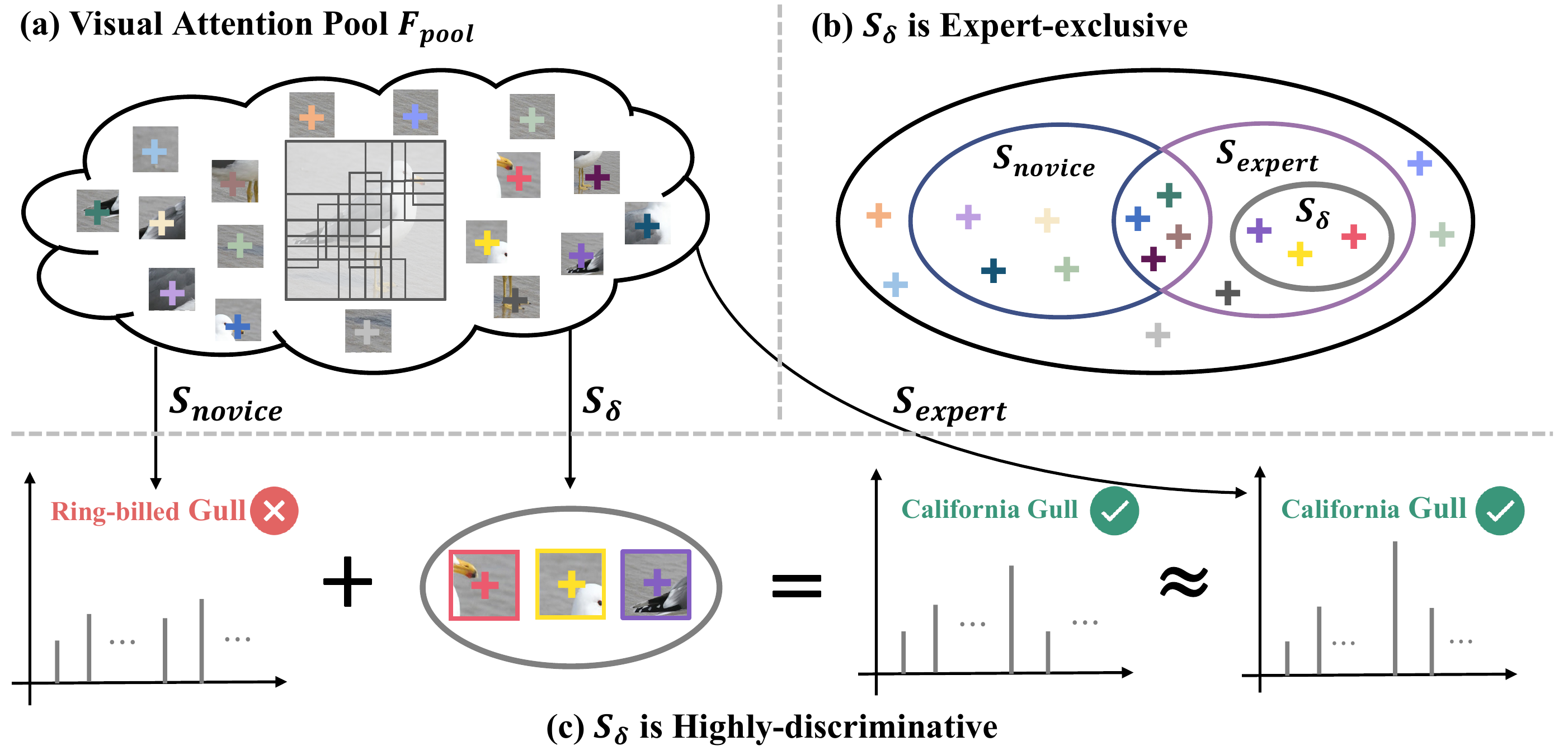}
\end{center}
\vspace{-0.3cm}
  \caption{Schematic illustration of how to obtain expert-exclusive but highly discriminative visual regions $S_{\delta}$ via our approach (Sec. \ref{sec:metho}). }
\label{fig:metho}
\end{figure*}

It follows that we define knowledge as highly descriminative visual regions that are exclusively attended by domain experts, \ie what parts of a photo experts focus on upon recognition. More specifically, we represent this expert attention as an optimal subset from a mixed pool of potentially discriminative regions (Fig.~\ref{fig:metho}(a)) that leads to maximum recognisability (Fig.~\ref{fig:metho}(c)). Our goal is then to eliminate non-expert ones that are shared between experts and novices so that expert-exclusive parts (knowledge) can be identified (Fig.~\ref{fig:metho}(b)). In accounting for novice knowledge, we show that fine-grained image caption works best amongst alternatives (\eg human annotated bubble regions (Sec.~\ref{sec:ablate}). Taken together, our technical solution is a multi-stage learning framework that (i) first conducts fine-grained representation learning in the visual domain, (ii) followed by associating human caption onto corresponding image regions, and (iii) distilling cross-modal attention differences to model expert-exclusive knowledge.

On measuring the efficacy of our bird guide (\ie knowledge learned from AI), we conduct a large-scale human study with a total of 15,000 trials on a fine-grained bird dataset \cite{wah2011caltech}. Results show, of the 2407 trials that participants initially failed fine-grained bird recognition, an average 53.39\% later successfully reverted their decisions, after being presented with our bird guide. We provide further analysis showing our approach is not constrained to work with bird species only, and improves conventional FGVC performance when exploited as a way to achieve explicit discriminative localisation.

\subsection{In Connection to Existing FGVC Works}
\label{sec:connection}

Our work is a general extension to the existing bulky literature of FGVC (a most recent survey at \cite{wei2021survey}), where we re-envisage the FGVC functionality from better label classification accuracy to that of providing useful knowledge for human consumption. It also brings out an important question of whether current FGVC methodologies in the traditional benchmarking sense have in fact learned any fine-grained knowledge -- the exact implication of that we however leave to future work.

From a knowledge dissemination perspective, the way how we reason and dissect a FGVC model also seems to resonate with recent literature on generating FGVC visual explanations \cite{selvaraju2017grad,chang2018explaining,wagner2019interpretable,chen2019looks,huang2020interpretable,huang2021stochastic}, especially at the first glimpse. A closer inspection however reveals the fundamentally different purposes they each serve. The goal of the existing works is on machine explainability, \ie looking into the pixels responsible for a model's decision and judging whether they align with human intuition or not (\eg by trying to make sense of visualised attention maps). We however take a human-centred view and only care if whatever extracted information can be instilled into our very brain as \textit{transferable knowledge}.

More precisely, existing works generally present a pixel selection function h($\cdot$) that either explains the decision of the black-box FGVC model in the form of post-hoc visualisation p(y$|$h(x)) \cite{selvaraju2017grad,chang2018explaining,wagner2019interpretable} or making FGVC an explainable model itself p(h(x)$|$y) \cite{chen2019looks,huang2020interpretable,huang2021stochastic}. As a result, h(x) inevitably contains many visual cues that most human novices can already perceive.
Our solution instead models p(y$|$h(x$\setminus$x\_human)), \ie we take into account of human (non-expert) prior knowledge of an image and exclude them from our knowledge base to ensure a well-defined expert representation. We verify the importance of doing so in Sec.~\ref{sec:expert-exclusive}.

\section{Methodology}\label{sec:metho}

In the traditional FGVC setting, given an image $x$ and its fine-grained label $y$ (\eg bird species name), a deep feature extractor $F(\cdot)$ will first process $x$ into a set of feature pool $F_{pool}= \{{f_1},{f_2},...,{f_N}\}$, representing a total size of $N$ visual features covering a diverse location and scale of visual regions. A classifier $Cls(\cdot)$ is then appended upon the rich visual information provided in $F_{pool}$ and optimised to predict $y$ under cross-entropy classification objective. The composition of $F(\cdot)$ and $Cls(\cdot)$ is therefore what we often regard as an AI-enabled domain expert.

Our goal is to extract highly discriminative visual regions that experts exclusively attend to in classifying a fine-grained image. Denoting the visual attentions of experts and novices as $S_{expert}$ and $S_{novice}$, this equates to learning an \textit{attention re-sampling operation} $S_{\delta}$ on $F_{pool}$ that can successfully bridge the gap between $S_{expert}$ and $S_{novice}$ in expert label prediction (Fig.~\ref{fig:metho}). Putting it formally:
\begin{equation}
\begin{aligned}
\label{eq:conceptual}
Cls(S_{expert}\odot F_{pool}) &\approx  Cls((S_{novice} + S_{\delta})\odot F_{pool}) \\
s.t.\; S_{\delta} & \subseteq S_{expert}\setminus(S_{expert} \cap S_{novice})\\
\end{aligned}
\end{equation}

\noindent We model $S_{expert}$, $S_{novice}$, $S_{\delta}$ as a learnable probability row vector ($\mathbb{R}^N$) in practice, \ie $\sum_{i=1}^{N}{s_*^i} = 1$. We shall now detail below how to obtain each component in Eq.~\ref{eq:conceptual}.

\subsection{Stage \RNum{1}: Visual Learning for $F_{pool},S_{expert},Cls(\cdot)$}

\keypoint{Obtaining $F_{pool}$.} Though obtaining $F_{pool}$ is not restricted to one specific method, it does need some careful consideration given $F_{pool}$ will be subjected to \textit{bi-modal} sampling from both $S_{expert}$ and $S_{novice}$. We find that the trivial workflow of learning $F_{pool}$, $S_{expert}$ and $Cls(\cdot)$ in end-to-end fashion will bias $F(\cdot)$ towards $S_{expert}$ and makes it incompatible to work with $S_{novice}$ later. For this, we propose a simple yet effective solution by decoupling the learning of $F_{pool}$ with that of $S_{expert}, Cls(\cdot)$. Specifically, given an image $x$, we divide it into $1\times1$, $2\times2$,..., and $k \times k$ uniform image blocks and use $F(\cdot)$ to extract feature for each local block to build our visual feature pool $F_{pool}$. We introduce an auxiliary classifier $Aux(\cdot)$ to guide the learning of $F_{pool}$ for ground-truth label predictions. $F_{pool}$ is then fixed with $Aux(\cdot)$ scrapped after this stage.

\keypoint{Obtaining $S_{expert}, Cls(\cdot)$.} We compute $S_{expert}$ by first conducting self-attention (SA)\footnote{We implement $\sa$ in the form of the popular Scaled Dot-Product Attention \cite{vaswani2017attention}: $SA(F_{pool}) = \softmax\left(\frac{Q_{pool}K_{pool}^T}{\sqrt{d_{pool}}}\right)V_{pool}$.} on $F_{pool}$ to better capture the long-term visual spatial dependency. We then append one fully-connected (FC) layer normalised with Softmax to simulate expert visual attention upon recognising a fine-grained object:
\begin{equation}
\begin{aligned}
\label{eq:expert}
&S_{expert} = \softmax(\fc(\sa(\Gamma(F_{pool})))) \\
\end{aligned}
\end{equation}
\noindent $\Gamma(\cdot)$ is a \texttt{stop\_gradient} operation that forbids gradient flowing through the variable it functions on, which we will apply throughout. 
Denoting $F_{expert}$ as $S_{expert}\odot F_{pool}$, we optimise $\{S_{expert}, Cls(\cdot)\}$ in the multi-label classification formulation:
\begin{equation}
\begin{aligned}
\label{eq:vision-obj}
&L_{vision} = \LCE(Cls(F_{expert}), y) 
\end{aligned}
\end{equation}

\subsection{Stage \RNum{2}: Visual Grounding for $S_{novice}$}

To bypass the otherwise fatal lack of human novice annotations on their perceivable visual regions of an image, we exploit the existing human fine-grained image caption dataset \cite{reed2016learning} to model $S_{novice}$. Given an image, ten single sentence visual descriptions are collected from different crowdsourced workers and we use their aggregate\footnote{To get image caption aggregate from different human visual descriptions, we use TextBlob \cite{textblob} to extract registered noun phrases from each human and combine them into one caption by eliminating the duplicates.} $c$ as a summary of the best possible visual perceptive zones from human novices. The question is how to ground human language input $c$ to the visual representation of $S_{novice}$?  We first process $c$ with an off-the-shelf pre-trained language model $\bert(\cdot)$ \cite{devlin2019bert} to get its semantic embedding $f_{c}$ and append a multi-layer perceptron $\Mlp(\cdot)$ aiming to project $f_{c}$ to an embedding space compatible with $F_{pool}$. $S_{novice}$ is then formulated as the broadcast element-wise cosine similarity between $\Mlp(f_{c})$ and $F_{pool}$:
\begin{equation}
\begin{aligned}
\label{eq:text-ground}
S_{novice} = \softmax(\cos(&\Mlp(\Gamma(f_c)),\Gamma(f_1)),\\
\cos(&\Mlp(\Gamma(f_c)),\Gamma(f_2)), ...,\\
\cos(&\Mlp(\Gamma(f_c)),\Gamma(f_N))) \\
\end{aligned}
\end{equation}
 
\noindent Since the role of $S_{novice}$ is to ensure human intentions expressed in language transfer visually, we require the training objective to maximise the cross-modal feature-wise mutual information $MI(\Mlp(f_c), F_{novice})$, where $F_{novice}=S_{novice} \odot F_{pool}$.

\keypoint{Noise contrastive learning.} Mutual information is notoriously intractable to optimise, where we resort to noise contrastive estimation \cite{gutmann2012noise} as a surrogate loss function. In particular, we implement it as InfoNCE \cite{oord2018representation} due to its wide adoption in the weakly-supervised visual grounding literature \cite{xiao2017weakly,gupta2020contrastive,wang2021improving,wang2021improving}. InfoNCE is manifested in the popular cross-entropy fashion and measures how well the model can classify one positive representation amongst a set of unrelated negative samples:
\begin{equation}
\begin{aligned}
\label{eq:ground-obj}
L_{ground} = \sum_{i=1}^{bs}-\log\frac{F_{novice,i} \cdot \Mlp(\Gamma(f_{c,i}))}{\sum_{j=1}^{bs}F_{novice,i} \cdot \Mlp(\Gamma(f_{c,j}))}
\end{aligned}
\end{equation}

\noindent with some slight abuse of notations, we use $F_{novice, *}$ and $f_{c,*}$ as the batch alternatives to $F_{novice}$ and $f_{c}$. $bs$ is the size of samples we use for contrastive learning with always $1$ positive and $bs-1$ negatives.

\subsection{Stage \RNum{3}: Knowledge Distillation for $S_{\delta}$}
\label{sec:kd}
Recall the two key traits of $S_{\delta}$ we defined conceptually. $S_{\delta}$ is first expert-exclusive visual attention on $F_{pool}$. This gives us the important prior information of the element-wise importance of $F_{pool}$ for $S_{\delta}$: $S_{\delta}$ attend to a subset of visual regions in $S_{expert}$ that is disjoint with $S_{novice}$, \ie $S_{\delta}$ corresponds to an attention re-sampling operation from the non-zero entry in $\max(S_{expert}-S_{novice},0)$. Similar to Eq.~\ref{eq:expert}, we model $S_{\delta}$ with feature-wise self-attention followed by one FC layer for output normalisation:
\begin{equation}
\begin{aligned}
\label{eq:delta}
S_{\delta} = &\softmax(\fc(\sa(\\
&\Gamma(F_{pool} \odot \max(S_{expert}-S_{novice},0)))))\\
\end{aligned}
\end{equation}

\noindent The second trait of $S_{\delta}$ is being highly discriminative that bridges the recognition gap between $S_{expert}$ and $S_{novice}$. Denoting the visual feature attended by $S_{\delta}$ as $F_{\delta}=F_{pool}\odot S_{\delta}$, 
we portray the learning as a process of knowledge distillation, where the student ($S_{\delta}$) tries to distil expert-exclusive knowledge from the teacher ($S_{expert}$):
\begin{equation}
\begin{aligned}
\label{eq:distil-obj}
    L_{distil} &= \kl(\\
    Cls&(\Gamma(F_{expert}/ t)) \;|| \;Cls((\Gamma(F_{novice}) + F_{\delta})/ t))
\end{aligned}
\end{equation}

\noindent where $\kl(\cdot)$ is the Kullback-Leibler divergence between two distributions and $t$ is the temperature hyperparameter \cite{hinton2015distilling, muller2019does,zhou2021rethinking} balancing the quality (sharpness) of the knowledge distilled, with smaller $t$ corresponding to fewer coverage of teacher's knowledge base and larger $t$ risking over smoothing out teacher's focus. We set $t=5$ throughout.

\noindent\textbf{Inference without reliance on $c$.}\; There is one shortcoming in Eq.~\ref{eq:distil-obj} when practically deployed: $S_{\delta}$ relies critically on the outcome of $S_{novice}$, which requires the fine-grained human language description of an image ($c$) from the user. We argue it's a big ask of user to offer the same level of descriptive comprehensiveness like those we use for training: ``This bird has a yellow, long, pointy beak, grayish feathers and grayish feathers, with white on the crown and black on the wingbars." We provide a simple solution to address this. A post-hoc approach is adopted that learns how to produce similar expert-exclusive discriminative visual attentions as with $S_{\delta}$ directly from $F_{pool}$:
\begin{equation}
\begin{aligned}
\label{eq:posthoc}
&\hat{S}_{\delta} = \softmax(\fc(\sa(\Gamma(F_{pool})))) \\
&\hat{F_{\delta}} = \hat{S}_{\delta} \odot F_{pool} \;\;\;\; L_{posthoc} = \mmd(\hat{F_{\delta}}, \Gamma(F_{\delta}))\\
\end{aligned}
\end{equation}

\noindent We choose Maximum Mean Discrepancy (MMD) as a discrepancy metric for its ability to distinguish between two distributions with finite samples:
\begin{equation}
\begin{aligned}
\label{eq:mmd}
&{\mmd}^{2}(\hat{F}_{\delta}, F_{\delta}) = \frac{1}{\binom{N}{2}}\sum_{i \neq i'}k(\hat{F}_{\delta, i},\hat{F}_{\delta,i'}) \\
&-\frac{1}{\binom{N}{2}}\sum_{i \neq j}k(\hat{F}_{\delta, i},F_{\delta, j}) +\frac{1}{\binom{N}{2}}\sum_{j \neq j'}k(F_{\delta, j},F_{\delta, j'})\\
\end{aligned}
\end{equation}

\noindent where $k(x,x')=exp(||x-x'||^2/\gamma)$ with $\gamma$ as a bandwidth hyperparameter. We show by replacing $S_{\delta}$ with $\hat{S}_{\delta}$ only brings marginal performance downgrading in our empirical evaluations.

\subsection{Incorporating $S_{\delta}$ as FGVC Booster}
\label{sec:fgvc-boost}
$S_{\delta}$ also provides an answer to the FGVC debate on what is the best way to achieve discriminative localisation \cite{zhang2014part,lin2015deep,deng2015leveraging,wang2018learning,ji2020attention}. Our speculation is that if $S_{\delta}$ has successfully encoded expert-exclusive visual attentions, visual regions $\delta$ corresponding to $S_{\delta}$ are then already locally discriminative with expert endorsement. There are of course many ways to embed $\delta$ as explicit localisation information into existing FGVC frameworks, of which we choose perhaps the most intuitive of seeing $\delta$ as another pixel space input together with the original image $x$. We find such simple solution suffices to bring improvements on top of many existing FGVC frameworks. To avoid ambiguity with existing notations, we abstract a FGVC solver into two parts of feature extractor $E(\cdot)$ and classifier $C(\cdot)$. We now formulate a new generic way of FGVC training with both $x$ and $\delta$ as input:
\begin{equation}
\begin{aligned}
\label{eq:notext}
L_{ce}= \LCE(C(E(x)+\sum_{i=1}^{N_{top}}E(\delta^i)),y)
\end{aligned}
\end{equation}

\noindent where in practice, we set $N_{top}=1$, \ie we only select the single most discriminative region in $\delta$ as one more image input for explicit localisation.

\section{Experiments}

Our main experiments are conducted on the CUB-Bird-200 dataset \cite{wah2011caltech}, which contains 11,877 images from a label categorisation of 200 bird species\footnote{Due to page space limit, we omit implementation details here and welcome the readers to find more in our codes upon paper publication.}. We first show how the learned knowledge in the visual form of expert-exclusive discriminative regions $S_{\delta}/\hat{S}_{\delta}$ can help people with divergent levels of bird expertise towards better recognising their once unrecognisable birds. We then confirm that $S_{\delta}$ is indeed attending to visual regions exclusive to domain experts and only with such type of visual feedback (\vs $\{S_{novice},S_{expert}\}$) can enable practically more interpretative and digestible knowledge to human participants. Given the lack of suitable baselines from existing FGVC research (elaborated in Sec.~\ref{sec:connection}), we move on to conduct ablation on our key technical choice of adopting image caption aggregate to represent the otherwise hard-to-quantify domain novice visual attentions. We wrap up the experiments by providing further analysis on $S_{\delta}$.

\begin{figure*}[!t]
  \centering
  \begin{subfigure}[t]{1.695in}
    \centering
    \includegraphics[width=1.695in]{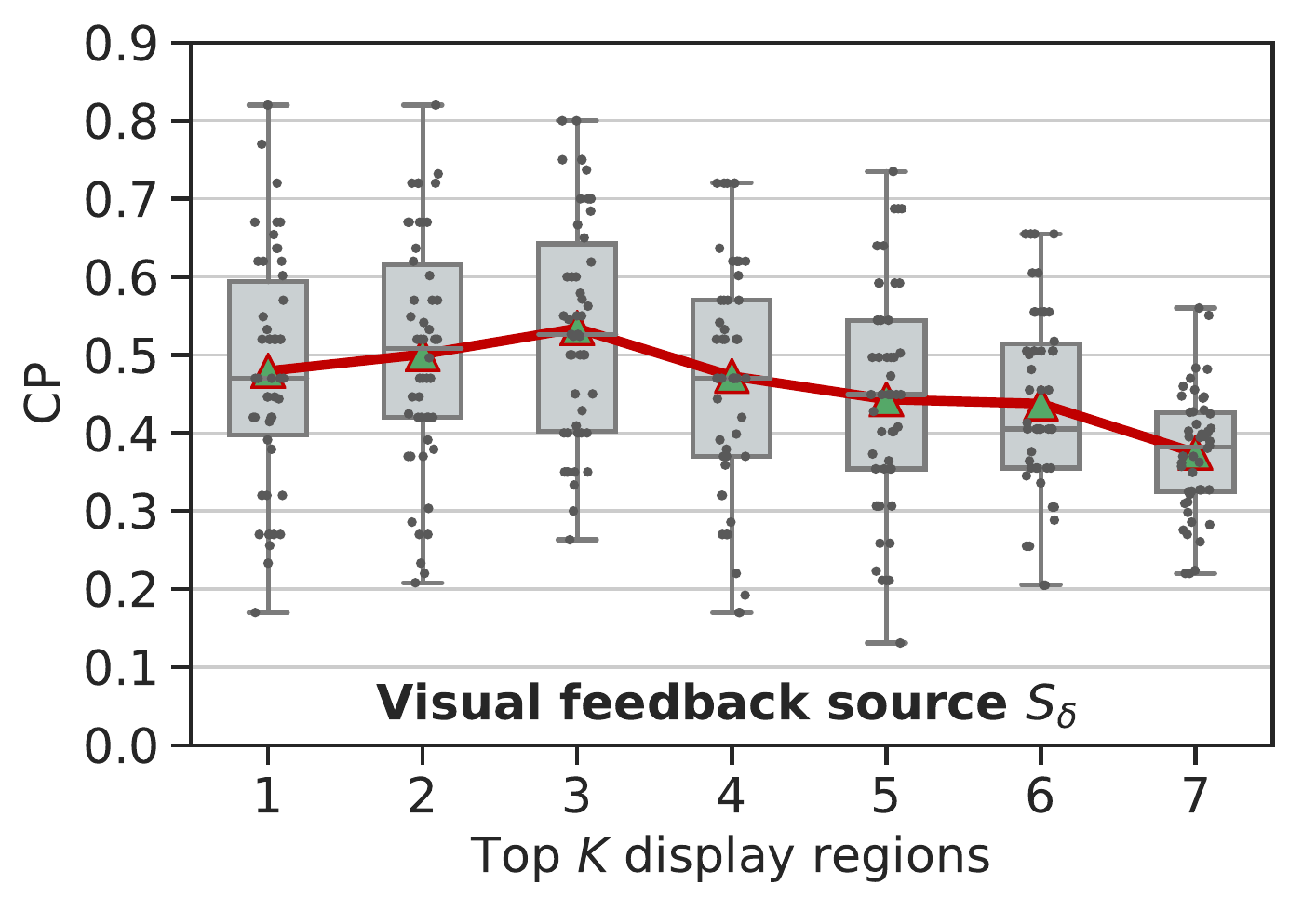}
 \end{subfigure}
  \begin{subfigure}[t]{1.695in}
    \centering
    \includegraphics[width=1.695in]{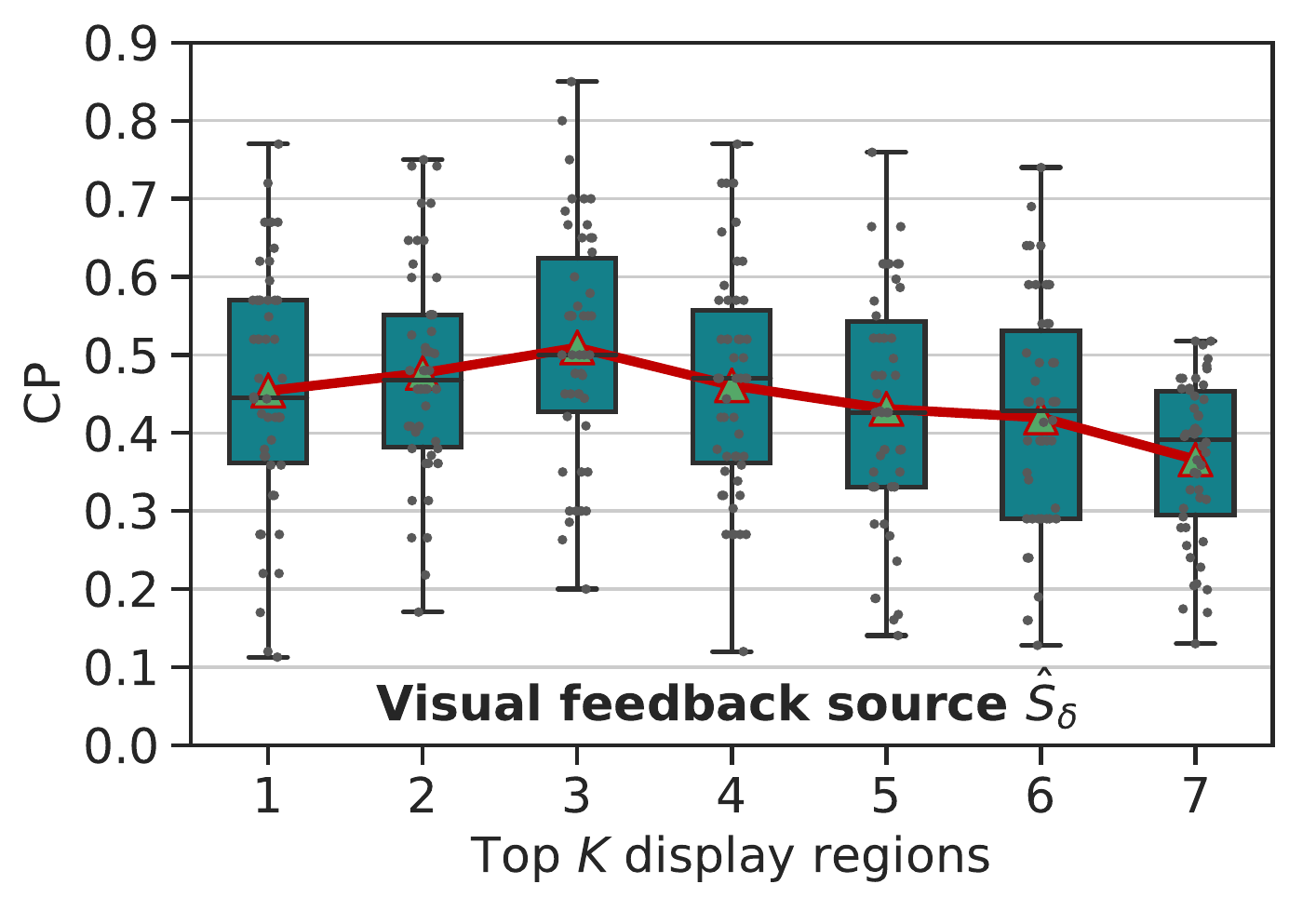}
 \end{subfigure}
  \begin{subfigure}[t]{1.695in}
    \centering
    \includegraphics[width=1.695in]{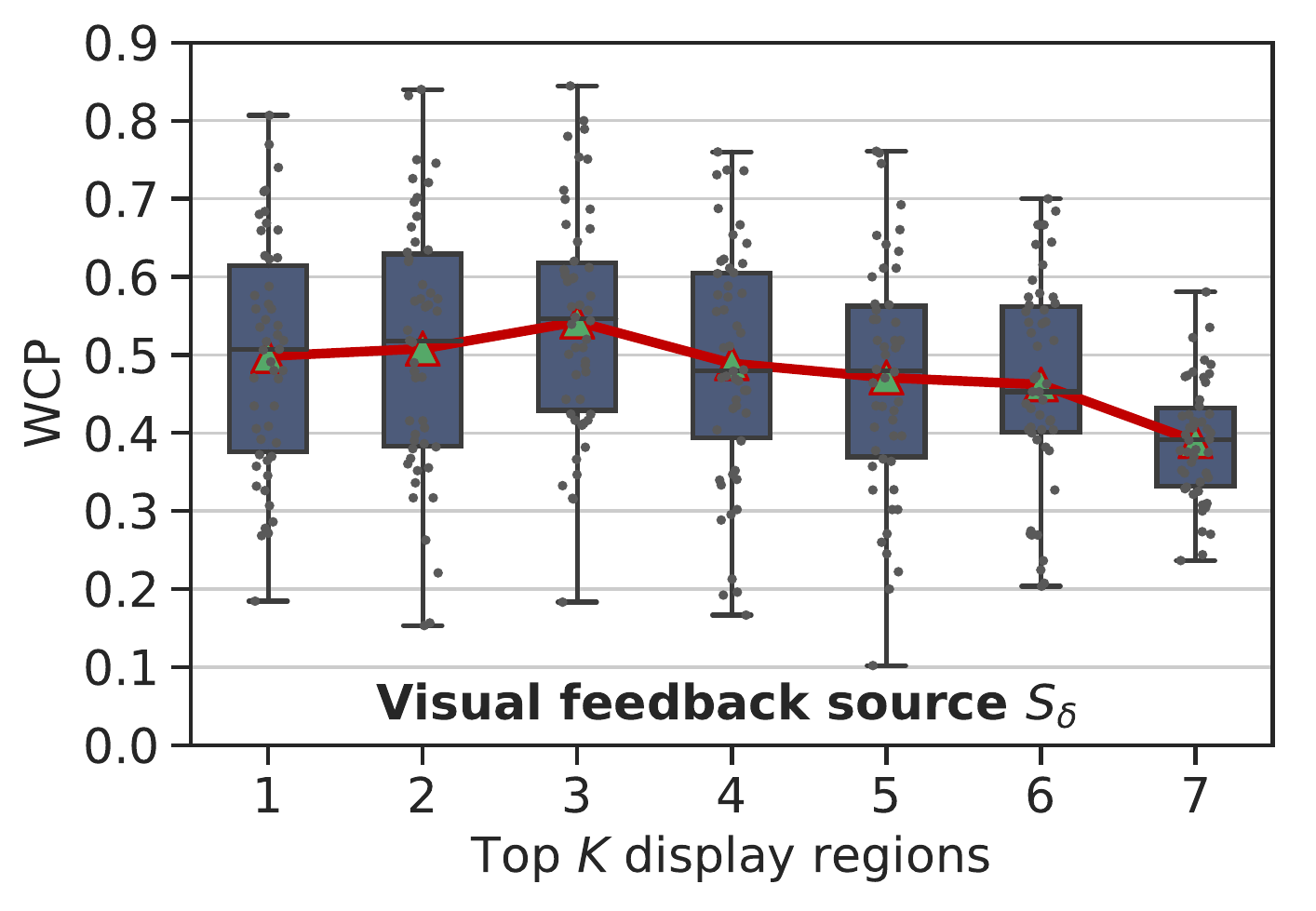}
 \end{subfigure}
  \begin{subfigure}[t]{1.695in}
    \centering
    \includegraphics[width=1.695in]{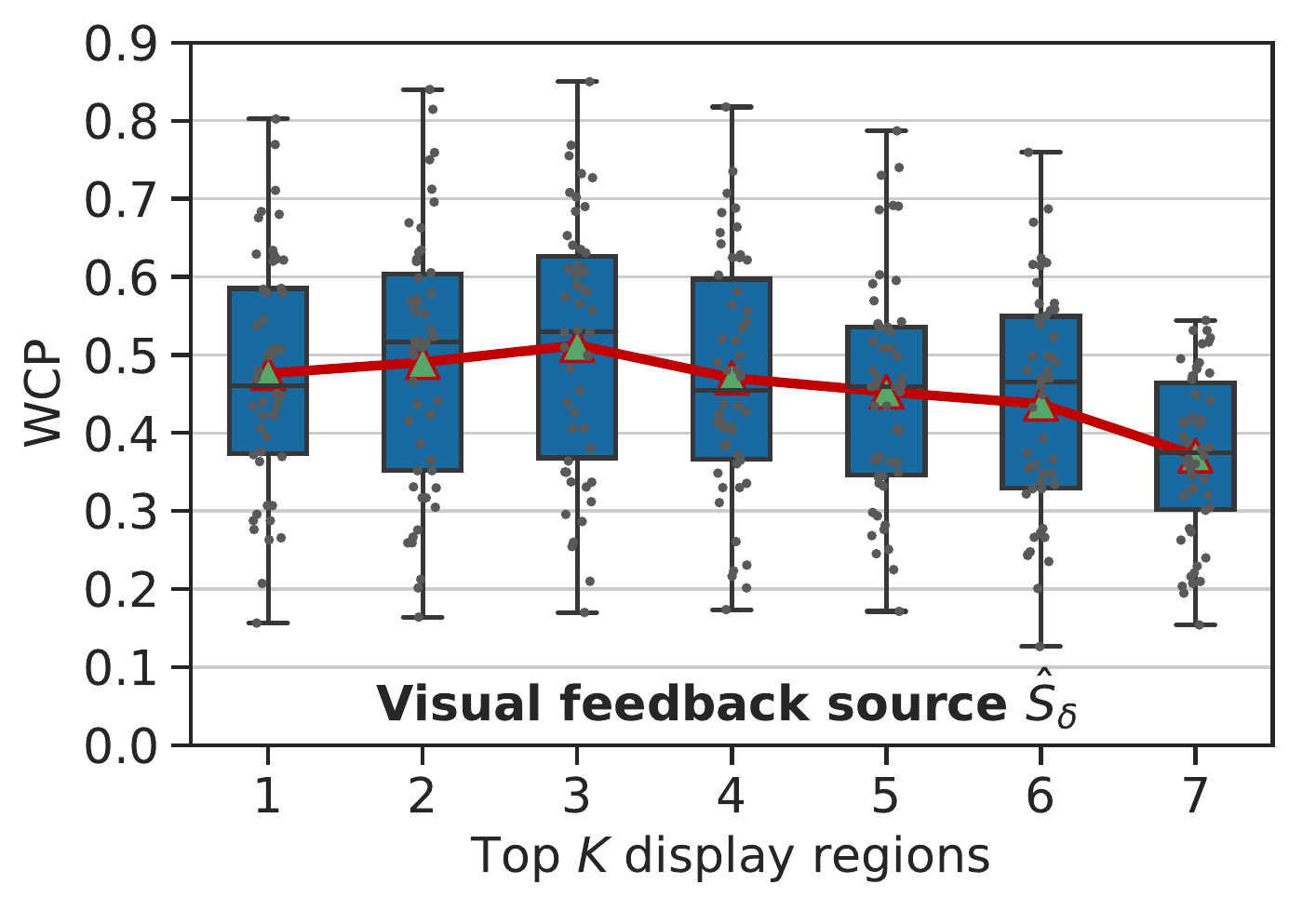}
 \end{subfigure}
  \begin{subfigure}[t]{6.85in}
    \centering
    \includegraphics[width=6.85in]{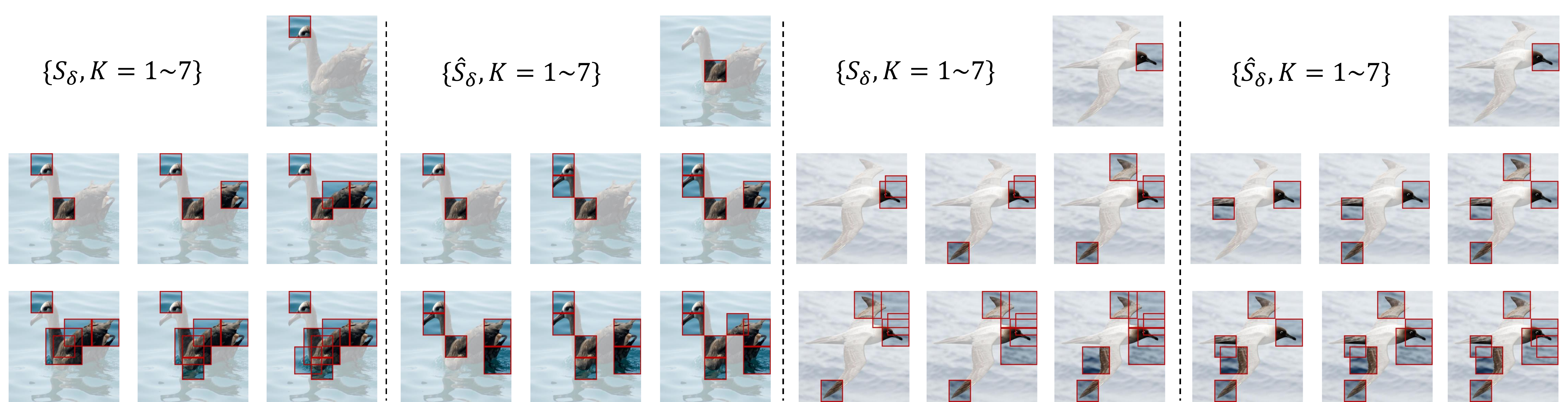}
 \end{subfigure}
  \caption{Top row: Box plot to demonstrate the efficacy of $S_{\delta}$/$\hat{S}_{\delta}$ in helping people reverse their failed decision for bird recognition. Green triangle represents the mean performance. Bottom two rows: Sample illustration of Top K visual regions $S_{\delta}$/$\hat{S}_{\delta}$ attend to.}
  \label{fig:main-exp}
\end{figure*}

\begin{figure}[t]
  \centering
  \includegraphics[width=\linewidth]{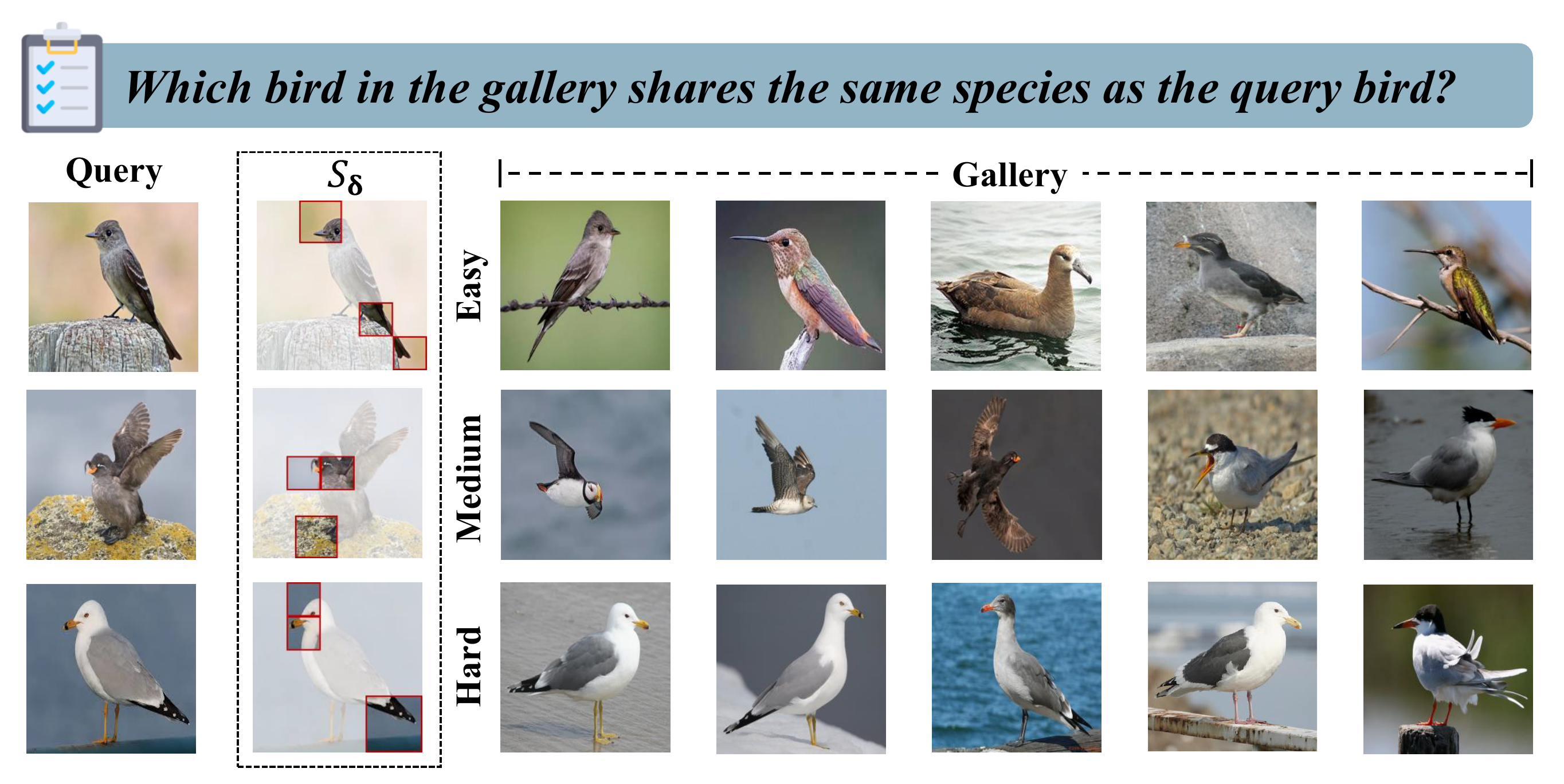}
  \vspace{-4mm}
  \caption{Sample questionnaire for measuring the efficacy of AI-empowered knowledge by simulating it as a book guide. In data and participant setup stage, $S_{\delta}$ is not shown.}
  \label{fig:questionaire}
\end{figure}

\keypoint{Data \& participant setup.} We recruit 200 participants across different ages, genders and education levels, where each of whom is expected to complete a questionnaire with 300 bird recognition tasks. In each task trial, the participant will be given a query bird image and five gallery bird images, and asked to select the \textit{only} image in the gallery that he/she believes to belong to the same bird sub-class with the query (Fig.~\ref{fig:questionaire}). The difficulty of the task then lies in the similarity level between the query and gallery images, where we define three challenge levels based on the biological bird stratification of Order-Family-Species: (i) \textit{Easy:} gallery samples are manifested in different bird orders. (ii) \textit{Medium:} gallery samples are from different bird families but all belonging to the same bird order. (iii) \textit{Hard:} gallery samples only differ in the finest species level, \ie they come from the same bird order and family. We assign different score for correct answer to question of different difficulty (0.5, 1, 1.5 point for easy, medium and hard respectively). This means when the 300 tasks are decomposed into three subsets of [90, 120, 90] for each challenge level (the setting we adopt), the full mark would be 300 points. We plot the normalised scoring histogram by counting the number of people falling in each of the 10 discretised bins and observe an intuitive Gaussian-like bell shape distribution. Since our goal is to simulate a study covering fairly for people with divergent levels of bird expertise, we form three representative population groups by randomly selecting [15,20,15] participants from the first three bins (worst scorers), the fourth to seventh bin (medium scorers) and the last three bins (best scorers) respectively. These 50 participants and their bird recognition cases are then setting our basis for the experiments later.

\begin{figure}[!t]
  \centering
  \begin{subfigure}[t]{0.519in}
    \centering
    \includegraphics[width=0.519in]{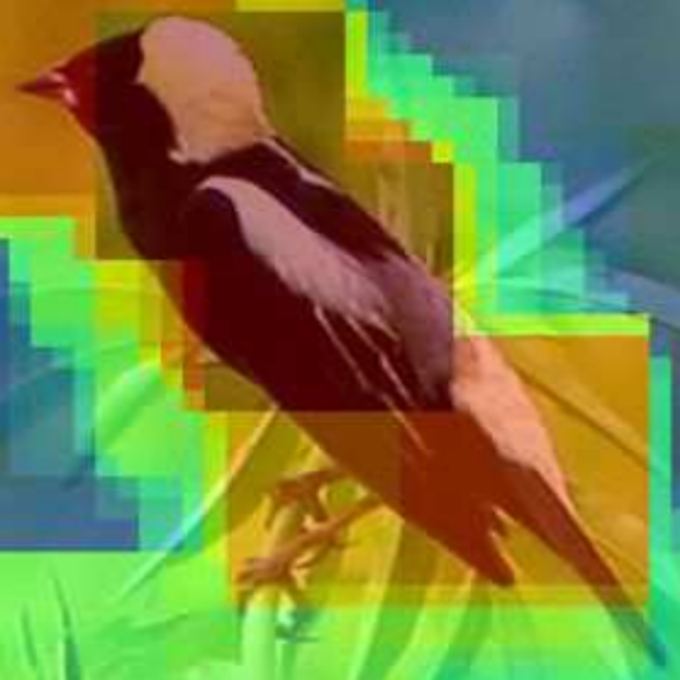}
    \caption*{\scriptsize{CVPR18~\cite{wang2018learning}}}
 \end{subfigure}
  \begin{subfigure}[t]{0.519in}
    \centering
    \includegraphics[width=0.519in]{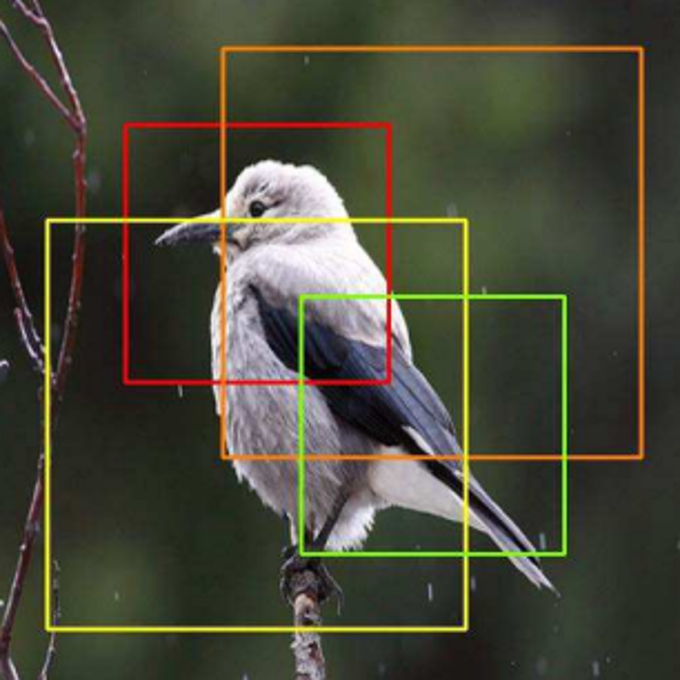}
    \caption*{\scriptsize{CVPR18~\cite{yang2018learning}}}
 \end{subfigure}
  \begin{subfigure}[t]{0.519in}
    \centering
    \includegraphics[width=0.519in]{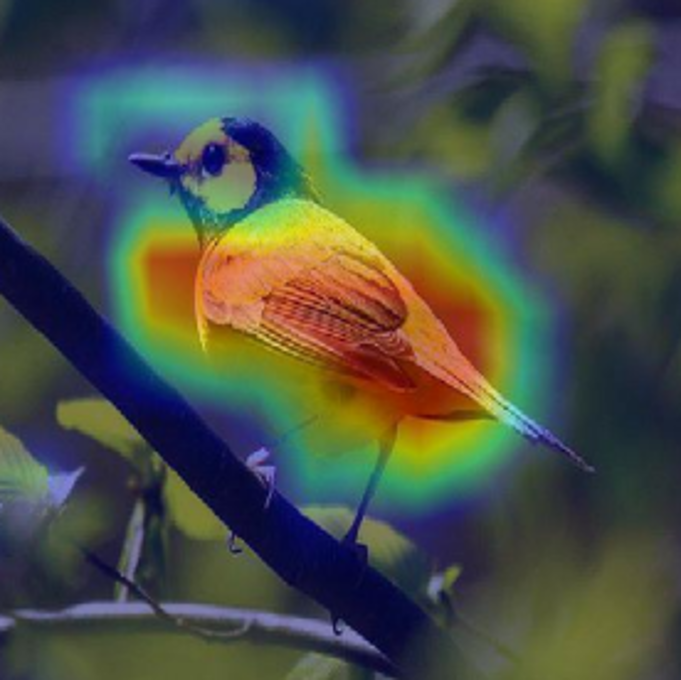}
    \caption*{\scriptsize{CVPR19~\cite{chen2019destruction}}}
 \end{subfigure}
  \begin{subfigure}[t]{0.519in}
    \centering
    \includegraphics[width=0.519in]{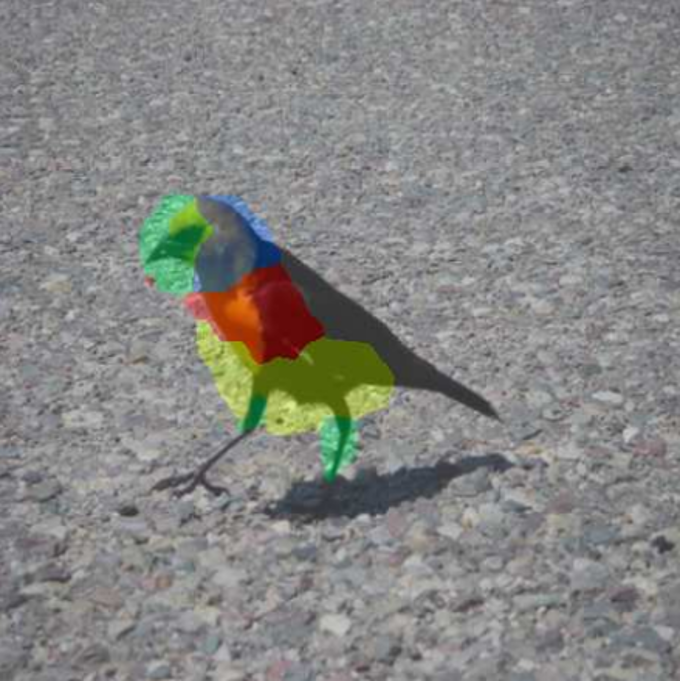}
    \caption*{\scriptsize{CVPR20~\cite{huang2020interpretable}}}
 \end{subfigure}
  \begin{subfigure}[t]{0.519in}
    \centering
    \includegraphics[width=0.519in]{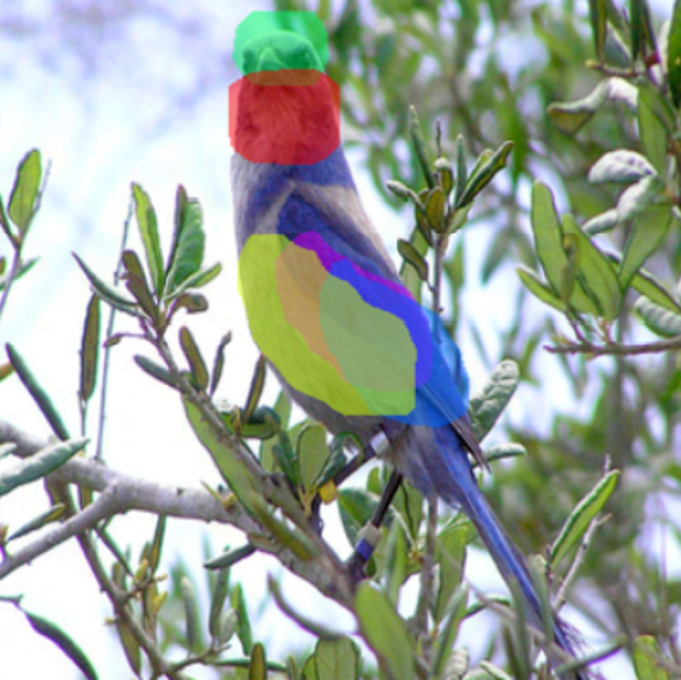}
    \caption*{\scriptsize{ICCV21~\cite{huang2021stochastic}}}
 \end{subfigure}
  \begin{subfigure}[t]{0.519in}
    \centering
    \includegraphics[width=0.519in]{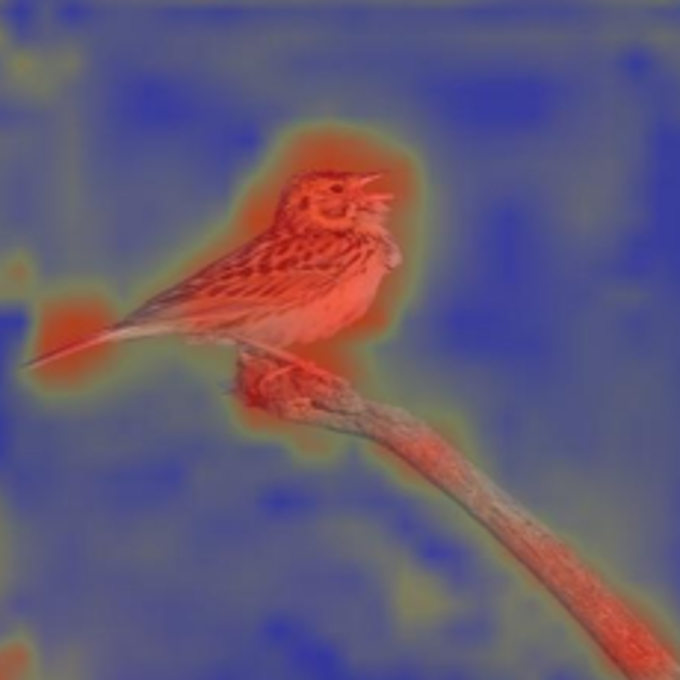}
    \caption*{\scriptsize{ICCV21~\cite{rao2021counterfactual}}}
 \end{subfigure}
  \caption{Visualisations of the typical supporting regions for the classifiers in the existing FGVC works. Examples shown are from direct copy-and-paste of the original paper.}
  \label{fig:visualization}
\end{figure}

\subsection{$S_{\delta}$ Is Your Expertly Curated Bird Guide}
\label{sec:exp-1}

\keypoint{Experimental method.} We now take the failed recognition cases from the 50 participants at the setup stage and examine the efficacy of $S_{\delta}$ on improving their recognisability. We also test the performance of $\hat{S}_{\delta}$, a practical alternative to $S_{\delta}$ when there is lack of fine-grained visual description of an image. We follow the similar ``query-gallery" experimental procedure, with just the difference that the query is now highlighted with the knowledge provided by $S_{\delta}$ (Fig.~\ref{fig:questionaire}). By paying extra attention to the AI-empowered knowledge, participants are required to re-make their decision of selecting the target image from the gallery that shares the same sub-class with the query. To mitigate the issue of participants intentionally altering their decision due to the awareness of their past failure or keeping their decisions unchanged because of behavioural inertia and memorisation, three strategies are taken to get around it: (i) we repeat the tests of their successful recognition cases and interleave them between failed ones, while not include these results in our statistics count; (ii) we enforce the participants to have at least have 24-hour time gap between undergoing any two different purposed experiments; (iii) We randomised the order of the individual task and its gallery images display. Our evaluation metric is twofold: average human correction percentage (\textbf{CP}) and average weighted human correction percentage (\textbf{WCP}). The former calculates the percentage of cases (2407 in total) that one human participant has successfully reverted their erroneous recognition under the guidance of $S_{\delta}$, where the latter corresponds to a weighted version that assesses the correction rate for cases in each challenge level first and weight it with the corresponding challenged point. Lastly, we rank the visual attentions of $S_{\delta}$ in descending order and always present the Top $K$ visual regions to human participants. We experiment with different $K$ values of [1,2,3,4,5,6,7] -- we find $K=$ 7 already brings notable degenerate performance in our pilot study as people tend to feel uncomfortable and fail to focus when faced with too many visual cues.

\noindent\textbf{Results.} We graphically describe CP and WCP for each human participant via the box plot of five-number quartile summary \cite{tukey1977exploratory} in Fig.~\ref{fig:main-exp}. Following observations can be made: (i) Under the metric of both CP and WCP, $S_{\delta}$ is able to provide the best performance when used to display its top 3 visual attentions with mean values, 53.39\% and 54.24\%. This provides compelling evidence that our learned $S_{\delta}$ is indeed extracting knowledge from an AI agent in a way that guides people towards better recognising a unknown bird. We further calculate the mean CP (mCP) and WCP (mWCP) values for people from three different scorer groups at setup stage, where the small differences among groups (52.63\%, 52.91\%, 54.91\%@mCP, 52.96\%, 53.88\%, 56.53\%@mWCP) confirm that $S_{\delta}$ is friendly and effective to users with divergent levels of bird expertise. (ii) Difference between $S_{\delta}$ and $\hat{S}_{\delta}$ is marginal in the eyes of human participants. This is an important message indicating that our framework can stand up to the fatal but common lack of per-image fine-grained language descriptions with little performance sacrifice. (iii) WCP values are slightly larger compared with those of CP. This is expected. Given $S_{\delta}$/$\hat{S}_{\delta}$ is designed to offer the most subtle expert-exclusive visual cues for successful fine-grained recognition, it naturally works better for solving harder cases with more bonus points (California Gull \vs Western Gull) compared with that for easier ones (Gull \vs Flamingo). (iv) There seems to exist a safety value (K$<$7) of how many visual regions to display and when the threshold is violated, humans start to show general failure in digesting the visual knowledge provided by $S_{\delta}$/$\hat{S}_{\delta}$. In line with psychological findings \cite{grant2003eye,roads2016using}, we ascribe such phenomenon to the fact that redundant visual distractors superimposed upon the most \textit{compact} visual highlights can be very detrimental for people to gain attentional expertise in practice. Interestingly, we also demonstrate some common visualisation results of existing FGVC works in Fig.~\ref{fig:visualization}. We can see how they generally cover the full attention map of a bird and correspond roughly to a K$=$7 scenario (or even worse!) under $S_{\delta}$/$\hat{S}_{\delta}$ -- indicating their natural unsuitability for human consumption as argued in Sec.~\ref{sec:connection}. We also consult our human participants on why they perform drastically poorer when K grows over a certain value and their response is unanimous: ``we don't know how to make sense from the knowledge manifested in crowded and cluttered visual regions.

\subsection{$S_{\delta}$ Works Because It Is Expert-Exclusive}
\label{sec:expert-exclusive}

In this section, we conduct deeper probe on $S_{\delta}$. Our goal is to show that $S_{\delta}$ is indeed distilling unique visual attentions from experts ($S_{expert}$) that are not shared by domain novices ($S_{novice}$), and this very property of $S_{\delta}$ consequently helps human participants to better recognise a bird. Below is detailed analysis.

\begin{figure}[t]
  \centering
\begin{subfigure}[t]{3.3in}
    \centering
    \includegraphics[width=3.3in]{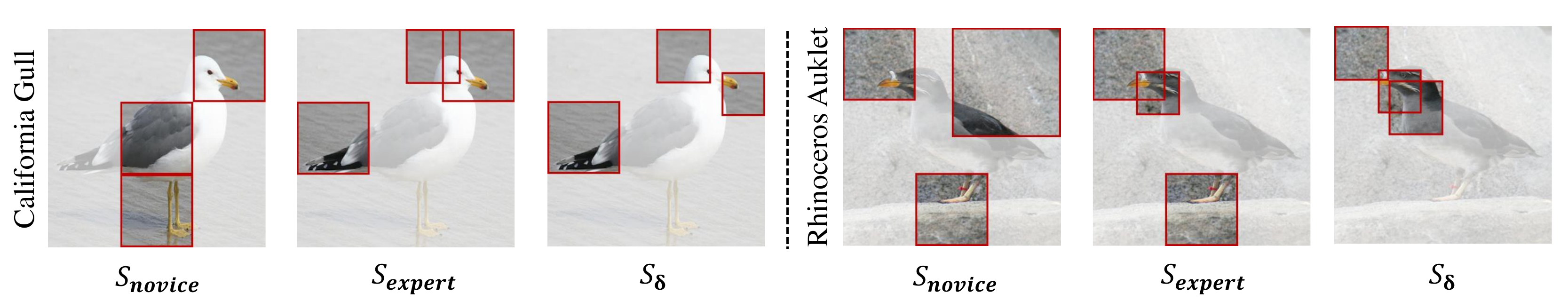}
    \caption{}
 \end{subfigure}
  \begin{subfigure}[t]{1.60in}
    \centering
    \includegraphics[width=1.60in]{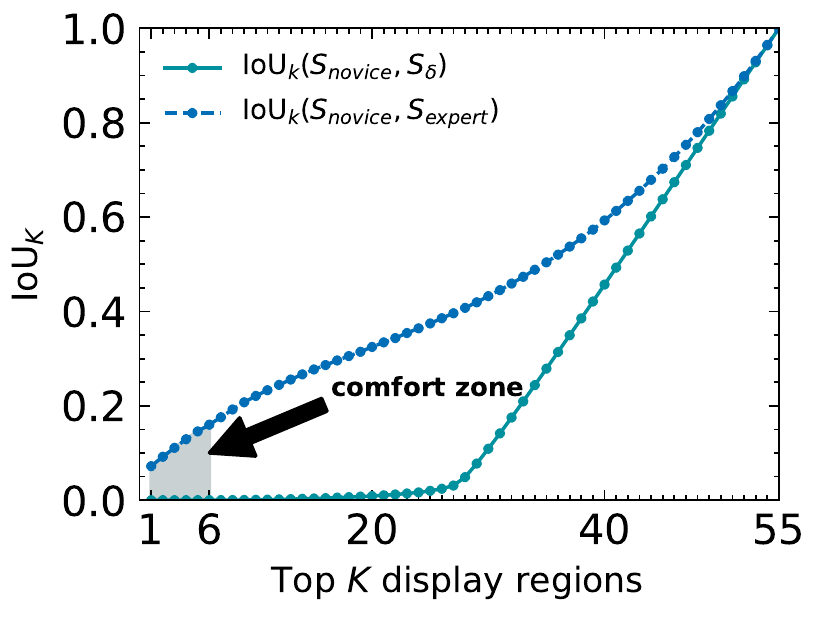}
    \caption{}
 \end{subfigure}
  \begin{subfigure}[t]{1.645in}
    \centering
    \includegraphics[width=1.645in]{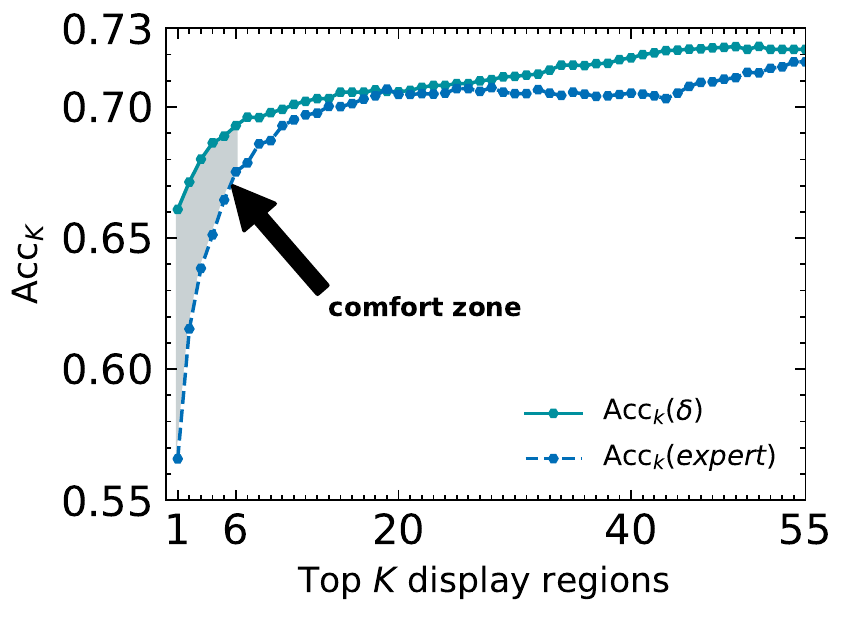}
    \caption{}
 \end{subfigure}
  \caption{(a): Exemplified comparisons among the Top 3 visual attentions encoded by $S_{novice}$, $S_{expert}$ and $S_{\delta}$. (b)(c): Understanding our learned $S_{\delta}$ from two different aspects. Comfort zone: maximum number of visual regions for display that we find humans can practically make sense of (K$<$7). More details in text.}
  \vspace{-0.3cm}
  \label{fig:analysis} 
\end{figure}

We first adopt Intersection over Union (IoU) to measure the correlation between the top $K$ rankings of two visual attention sequences -- if $\boldsymbol{\iou_{K}(S_{novice},S_{expert})}$ is significantly larger than $\boldsymbol{\iou_{K}(S_{novice},S_{\delta})}$ before $K$ grows impractically large, we know $S_{\delta}$ has successfully extracted the exclusive parts from $S_{expert}$. Results in 
Fig. \ref{fig:analysis}(a)(b) confirm that $S_{\delta}$ indeed shares negligible ($\le 0.01$) attentional overlap with $S_{novice}$ for K up to 20, in a stark contrast with the strong correlated interplay between $S_{expert}$ and $S_{novice}$. To shed further light on the importance of refining $S_{expert}$ to $S_{\delta}$ and its practical meaning as a form of knowledge to human participants, we calculate the expert label prediction accuracy $\boldsymbol{\acc_K(\delta)}$ and $\boldsymbol{\acc_K(expert)}$\footnote{To obtain $\acc_K(\delta)$ ($\acc_K(expert)$) on different $K$ values, we first work out normalised mean feature representation $\frac{1}{2}(f(F_{novice})+ \sum_{i=1}^{K}\lambda S_{\delta}^iF_{\delta}^i)$, s.t. $\lambda = 1/\sum_{i=1}^{K}S_{\delta}^i$ before feeding it into classifier.} with the combined visual cues from $\{S_{novice},\sum_{i=1}^{K} S_{\delta}^i\}$ and $\{S_{novice},\sum_{i=1}^{K}S_{expert}^i\}$ respectively. Our intuition is that if $S_{\delta}$ provides practically more useful complementary knowledge to what human already knows, $\acc_K(\delta)$ should reach to a considerably satisfying performance at a much smaller K than that of $\acc_K(expert)$. In other words, knowledge encoded in $S_{\delta}$ is more condensed and effective for human to digest because of its nature of expert-exclusive. Fig.~\ref{fig:analysis}(c) shows this is exactly the case where $\approx$ 91.40\% of label prediction performance is retained with only \textit{one} best visual attention in $S_{\delta}$ and up to $\approx$ 94.05\% when K$=$3. We also repeat the ``query-gallery" experiment in Sec.~\ref{sec:exp-1} and aim to figure out to what extent can $S_{novice}$ and $S_{expert}$ improve people's bird recognisability in a human study. By examining their performance under both mCP and mWCP and comparing them with $S_{\delta}$ (40.02\% and 47.05\% \vs 53.39\% @mCP, 39.56\% and 45.51\% \vs 54.24\% @mWCP), we can fairly conclude that the hypothesis of fine-grained visual knowledge being expert-exclusive does matter for practically more effective human consumption.

\subsection{Good $S_{\delta}$ Solver Needs Human Language Input}
\label{sec:ablate}
A critical part of our framework at \textit{design-level} is how to define and quantify $S_{novice}$ with data at hand. Given the rich visual elements of an image and the subjective nature of human vision on their relative importance, deciding the best form of representing domain novice visual attentions becomes indeed an art of choice. Our proposed method advocates the use of human fine-grained caption aggregate of an image for learning $S_{novice}$, where we compare it with several competitors below.

\begin{table}[t]
  \centering
\begin{adjustbox}{width=\linewidth,center}
       \Huge   
    \begin{tabular}{cccc}
    \toprule
\textbf{Raw Image} & \textbf{Annotated Bubbles} & \textbf{Human Drawing} & \textbf{Caption Aggregate}\\
    \midrule
    \midrule
    \makecell*[c]{\includegraphics[width=0.8\linewidth]{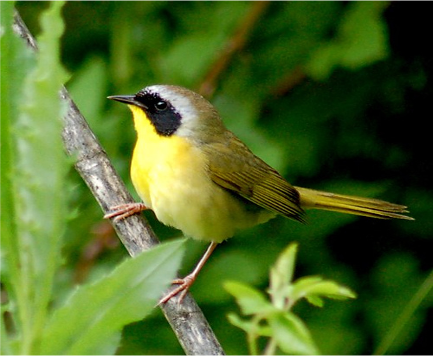}} & 
    \makecell*[c]{\includegraphics[width=0.8\linewidth]{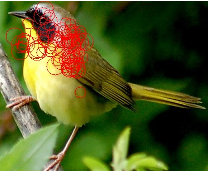}} & 
    \makecell*[c]{\includegraphics[width=0.8\linewidth]{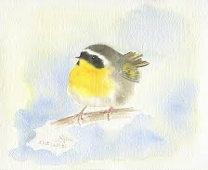}} & 
      \makecell*[c]{This bird has a yellow-breasted, \\
    a black cheek patch, \\
    and a white superciliary...} \\
    \midrule\noalign{\smallskip}
    \makecell*[c]{\includegraphics[width=0.8\linewidth]{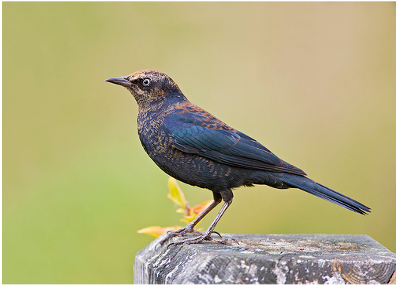}} & 
    \makecell*[c]{\includegraphics[width=0.8\linewidth]{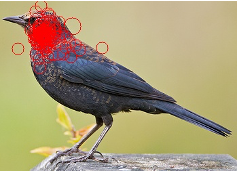}} & 
    \makecell*[c]{\includegraphics[width=0.8\linewidth]{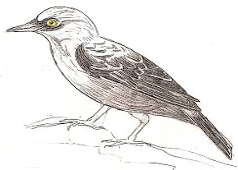}} & 
      \makecell*[c]{This bird has long legs, \\
    short tan wings, a small bill, \\
    and grayish-brown belly feathers...} \\
    \bottomrule
    \end{tabular}%
\end{adjustbox}
\vspace{-0.1cm}
\caption{Human annotated bubbles and drawings as alternatives of caption aggregate (Ours) to representing $S_{novice}$.}
  \label{tab:bubble}%
\end{table}%

\begin{table}[t]
    \centering
    \resizebox{\linewidth}{!}{
    \begin{tabular}{@{}lcccc@{}}
    \hline
    -- &\textbf{Ours} &Bubble \cite{deng2015leveraging} &Drawing \cite{wang2020progressive} &Beginner \cite{chang2021your} \\
    \cdashline{1-5}
    mCP & $53.39$ & $48.40$ & $49.07$ & $50.04$ \\
    \cdashline{1-5}
    mWCP & $54.24$ & $46.98$ & $49.69$ & $51.68$ \\
    \hline
    \end{tabular}}
    \vspace{-0.1cm}
    \caption{Performance comparisons (\%) between different realisations of $S_{novice}$ defined in Sec. \ref{sec:ablate}.}
    \label{tab:ablate}
    \end{table}

\begin{figure}[t]
\centering
\includegraphics[width=\linewidth]{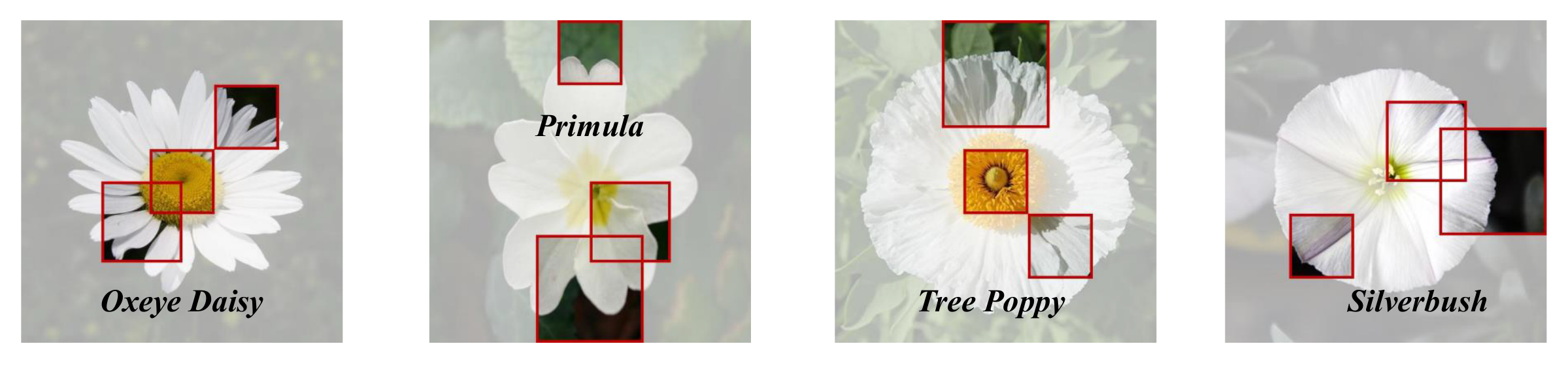}
\caption{Typical Top 3 visual attentions of $S_{\delta}$ when trained on fine-grained flower dataset.}
  \label{fig:flower}
\end{figure}

\begin{table}[t]
\centering
\resizebox{\linewidth}{!}{
\begin{tabular}{@{}l|cc|cc@{}}
\hline
\multirow{2}[2]{*}{Method} & \multicolumn{2}{c|}{CUB-Bird-200} & \multicolumn{2}{c}{Oxford-Flower-102} \\
\cdashline{2-5} & \multicolumn{1}{c}{Baseline$^{\ddagger}$} & \multicolumn{1}{c|}{\textbf{Ours}} & \multicolumn{1}{c}{Baseline$^{\ddagger}$} & \multicolumn{1}{c}{\textbf{Ours}} \\
\hline\hline
B-CNN (ICCV15 ~\cite{lin2015bilinear}) &  $87.16$     &     $\bf{87.78}$  & $96.77$      & $\bf{97.35}$  \\
NTS (ECCV18~\cite{yang2018learning})   &  $87.02$     & $\bf{87.54}$      &  $95.84$     & $\bf{96.51}$ \\
PC (ECCV18~\cite{dubey2018pairwise}    &    $86.71$   & $\bf{87.34}$      & $97.03$      & $\bf{97.42}$ \\
DCL (CVPR19~\cite{chen2019destruction})   &  $87.31$     &$\bf{87.86}$       &    $96.49$   &  $\bf{97.12}$\\
CrossX (ICCV19~\cite{luo2019cross})  &  $87.36$     & $\bf{87.84}$       &  $97.06$     & $\bf{97.36}$ \\
PMG (ECCV20~\cite{du2020fine})  &  $89.62$     & $\bf{89.74}$      &     ${98.02}$  & $\bf{98.21}$ \\
DeiT-B (ICML2021~\cite{touvron2021training})  &   $90.04$  &   $\bf{90.15}$  & $98.16$    & $\bf{98.28}$ \\
ViT-B-16 (ICLR2021~\cite{dosovitskiy2020image})  &   $90.33$ & $\bf{90.42}$    & $99.04$  & $\bf{99.17}$  \\
\hdashline
CVL$^{\S}$ (CVPR17~\cite{he2017fine})  &   $86.74$    &  $\bf{87.02}$    &    $96.87$   & $\bf{97.23}$ \\
PMA$^{\S}$ (TIP20~\cite{song2020bi})  &    $88.70$   &   $\bf{88.92}$    &   $97.12$    & $\bf{97.65}$ \\
\hline
\end{tabular}}
\vspace{-0.2cm}
\caption{$S_{\delta}$ improves existing FGVC methods when exploited for providing localisation information. $^{\ddagger}$: results obtained using publicly released code. $^{\S}$: methods that use fine-grained text descriptions as side information. }
\label{tab:fgvc}
\end{table}

\keypoint{Competitors.} We include three competitors for different conceptualisations of $S_{novice}$: (1) \textit{Discriminative bird bubbles}: These annotated bird circular regions (Tab.~\ref{tab:bubble}), namely ``bubbles" \cite{deng2015leveraging}, are collected via a novel online game aiming to reveal the most discriminative parts of a bird image. We aggregate the available bubbles of an image from multiple players and use their mean ImageNet pre-trained feature representation to learn $S_{novice}$. (2) \textit{Human bird drawings}: CUB-200-Painting \cite{wang2020progressive} is an extension of CUB-200-2011 bird dataset, which contains diverse human drawing forms (Tab.~\ref{tab:bubble}) aiming to visually interpret a fine-grained bird species, including watercolors, oil paintings, sketches and cartoons. We aggregate the human drawings under one bird species and use their mean ImageNet pre-trained feature representation to learn $S_{novice}$. (3) \textit{Junior bird expert}: We also model $S_{novice}$ as a beginner-level bird specialist that can differentiate between 13 bird subclasses at order level \cite{chang2021your} -- instead of the finer recognition of 200 subclasses at species level required towards a bird expert. For this, we train a 13-way classification model and adopt it (like ImageNet pre-trained feature) to learn $S_{novice}$.

\keypoint{Results.} We follow the same ``query-gallery'' experimental procedures in Sec. \ref{sec:exp-1} to evaluate the knowledge efficacy of $S_{\delta}$ provided by the different realisations of $S_{novice}$ as described above. We report the result in Tab. \ref{tab:ablate} and confirm the significance of our technical choice of using fine-grained caption aggregate to represent what domain novice can perceive from an image. Interestingly, the worst choice of $S_{novice}$ (Bubble) still outperforms $S_{expert}$ and $S_{novice}$ (48.40\% \vs 47.05\% and 40.02\% @mCP, 46.98\% \vs 45.51\% and 39.56\% @mWCP), stressing again the importance of our expert-exclusive modelling.

\subsection{Further Analysis On $S_{\delta}$}

\keypoint{$S_{\delta}$ works beyond birds.} We repeat the learning process on the fine-grained flower dataset \cite{nilsback2008automated} and conduct the same human study pipeline as with birds. The mCP performance of 69.39\% when displaying Top 3 visual region from $S_{\delta}$ confirms its efficacy as digestible knowledge in helping human participants to better recognise an unknown flower type. We show some visualisations of $S_{\delta}$ in Fig.~\ref{fig:flower}. 

\keypoint{$S_{\delta}$ improves FGVC performance.} We embed $S_{\delta}$ into existing FGVC frameworks aiming to help model better localise to the most discriminative visual regions (detail in Sec.~\ref{sec:fgvc-boost}). We confirm in Tab.~\ref{tab:fgvc} that $S_{\delta}$ is indeed a promising universal FGVC booster regardless of the base models built upon. Notably, our result also improves over CVL and PMA, two methods that have already specifically hedged their bets on the fine-grained textural information for more discriminative attention modelling. 

\section{Conclusion}

Results from a human study indicate that our method is able to obtain useful human-digestible knowledge from a FGVC model that significantly improves participants' ability on distinguishing between fine-grained objects. This is made possible by first representing knowledge as those visual regions attended exclusively by domain experts and managing to model it with a multi-stage cross-modal learning framework. By taking a first and firm step towards enabling a FGVC model as a human knowledge provider, this work could apply to a wide range of end-user applications that require fine-grained recognition outputs to be accessible for human consumption. Last but not least, we hope to have caused a stir and help to trigger potential discussions on how to make a heavily-invested AI expert system doing more good for all.

{\small
\bibliographystyle{ieee_fullname}
\bibliography{main}
}

\end{document}